\documentclass[letterpaper, 10 pt, conference]{ieeeconf}
\IEEEoverridecommandlockouts 
\let\labelindent\relax
\usepackage{hyperref, url, amsmath, amssymb, booktabs, graphicx, bm, color, algorithm, tabularx, enumitem, multirow, threeparttable, flushend}
\usepackage[noend]{algpseudocode}
\graphicspath{{images/}}

\newcolumntype{Y}{>{\centering\arraybackslash}X}

\makeatletter

\newcommand{\Rmnum}[1]{\expandafter\@slowromancap\romannumeral #1@}
\makeatother

\title{\LARGE \bf Characterization of a Meso-Scale Wearable Robot \\ for Bathing Assistance}

\author{Fukang Liu, Vaidehi Patil, Zackory Erickson, and Zeynep Temel
\thanks{Fukang Liu is with the Mechanical Engineering Department and Vaidehi Patil, Zackory Erickson, and Zeynep Temel are with the Robotics Institute, Carnegie Mellon University, Pittsburgh, PA, USA.}%
}

\begin{document}

\maketitle
\thispagestyle{empty}
\pagestyle{empty}

\begin{abstract}
Robotic bathing assistance has long been considered an important and practical task in healthcare. Yet, achieving flexible and efficient cleaning tasks on the human body is challenging, since washing the body involves direct human-robot physical contact and simple, safe, and effective devices are needed for bathing and hygiene. In this paper, we present a meso-scale wearable robot that can locomote along the human body to provide bathing and skin care assistance. We evaluated the cleaning performance of the robot system under different scenarios. The experiments on the pipe show that the robot can achieve cleaning percentage over $92$\% with two types of stretchable fabrics. The robot removed most of the debris with average values of $94$\% on a human arm and $93$\% on a manikin torso. The results demonstrate that the robot exhibits high performance in cleaning tasks.
\end{abstract}


\section{Introduction}
\label{sec:intro}

Robotic caregivers that assist with self-care tasks, such as bathing and skin care assistance, present an opportunity to positively impact the lives of millions of older adults and people with physical impairments. Recent estimates by the World Health Organization (WHO) have found that over 100 million adults over the age of $60$ are care-dependent and rely on assistance to complete basic tasks needed for daily living~\cite{world2017integrated}. In addition, global populations are aging and it is projected that the proportion of people over the age of $60$ will surpass $22\%$ of the total global population by $2050$~\cite{united2019world}. Compounding this issue, many nations are already experiencing shortages of professional human caregivers, which places a heavy burden on informal caregivers, whom are often family members and friends~\cite{GCoABuilding}. Among activities of daily living (ADLs), repeated studies have found that bathing is among the first tasks that adults may require assistance with and is a strong predictor of disability in other ADLs~\cite{gill2006epidemiology, ahluwalia2010perspectives, theou2012disability}.

 
Current robotic caregiving platforms often consist of a fixed-base or mobile manipulator that physically interacts with people~\cite{zlatintsi2020support,gao2016iterative,park2018multimodal,luo2017robot}, or mobility devices, such as robotic walkers and wheelchairs~\cite{yamazaki2012home, yang2014home}. 
Recent advances in robotic bathing assistance have also introduced robot manipulators that use flat rigid bathing tools to contact and clean a small region of the surface area of a human limb~\cite{Erickson2019dressing, 5649101, erickson2020assistive}. Yet, such approaches require robots to make multiple passes using complex trajectories around the human body to provide assistance. 
In comparison to these approaches, we propose a new platform and perspective to robotic caregiving in which a robot directly attaches to and locomotes along the human body to provide care.

\begin{figure}
\centering
\includegraphics[width=0.45\textwidth, trim={0cm 0cm 0cm 0cm}, clip]{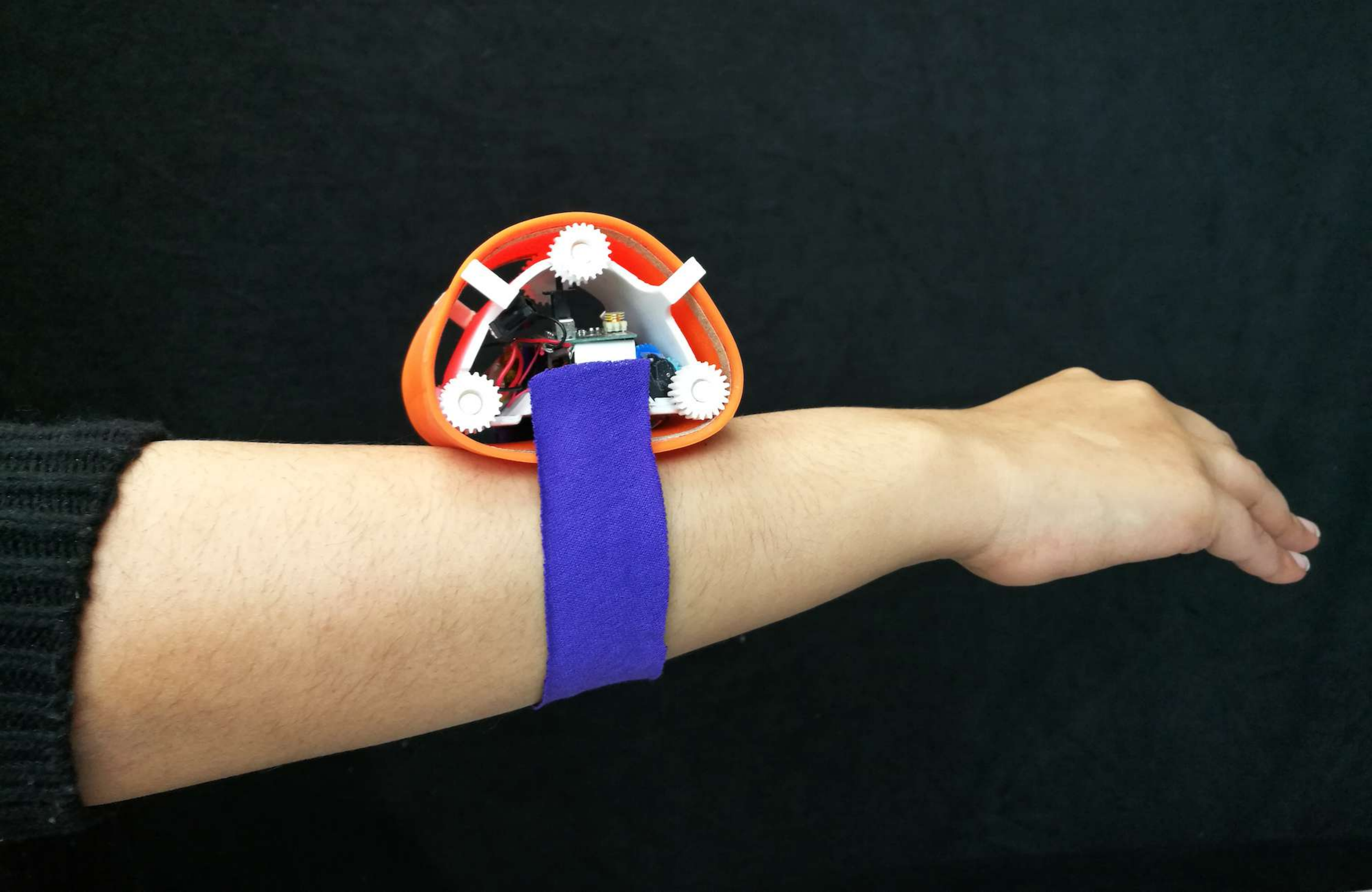}
\caption{\label{fig:ScrunchieBot} Autonomous meso-scale cleaning robot moving on a human arm. }
\label{ScrunchieBot}
\end{figure}

Here, we present an wearable robot that locomotes along cylindrical structures to provide cleaning, which has potential to be used in bathing and skin care assistance (Fig.~\ref{fig:ScrunchieBot}). With a characteristic length of 72 mm and weight of 63.4 g, our robot is comparable to existing wearable measuring devices, e.g., wrist blood pressure monitors usually have characteristic length and weight of $\sim10$ cm and $\sim125$ g, respectively~\cite{andmedical}. In addition, the soft surface of the robot and the cloth attachment enable a safe physical interaction with human skin. The mobility of the robot is based on track-drive locomotion to provide continuity of contact between the robot and the surface it is moving on. We show the cleaning performance of four different fabric attachments with our experiments on a PVC pipe that is soiled by colored powder. The results show that our robot has more than $95\%$ success rate in cleaning cylindrical pipes. In addition, we present the cleaning performance of the system on a human arm and medical manikin torso. Our contributions are as follows: 
\begin{itemize}
    \item We present a novel approach on wearable caregiving robots by introducing a meso-scale autonomous robot that moves along cylindrical objects and has potential to provide assistance with bathing and skin care.
\item We evaluate the performance of our robot in a series of controlled experiments. We focus on the physical parameters of fabric attachments to understand the optimum robot requirements that maximizes the cleaning performance. 
\item We present the cleaning performance of our robot on the human arm of one of the experimenters. Our robot achieves limb cleaning tasks reliably and effectively. 
\item We present results of our preliminary experiments that shows the cleaning ability of our robot on the torso of a human manikin. Results demonstrates effectiveness and potential of the robot for body cleaning tasks.
\end{itemize}

\section{Related Work}
\label{sec:related_work}

\subsection{Miniature Robots that Move on the Body}
Increasing interest in assistive technologies for health monitoring and patient care has led to the development of small-scale, soft sensing and actuation systems for wearable devices \cite{sanchez2021soft}. To date, researchers have developed several small-scale robots that can hold on to and move along the human body. Some of these studies focus on moving on clothing, such as the cm-scale companion robot developed by Liu and Wu~\textit{et al.}~\cite{liu2012system, liu2012path, wu2017movement,geng2018clothbot_beta}, where they demonstrated autonomous navigation on flexible materials. Saga~\textit{et al.}~\cite{saga2014daily} also proposed a small wearable companion robot that can move on a dedicated belt with fixed rails. Another robot that can move on clothing was proposed by Dementyev~\textit{et al.}~\cite{dementyev2016rovables}, which is smaller and lighter than previous examples ($40\times26\times26$ (mm), 20 g). This robot, Rovables, was used for interactive clothing and haptic feedback.

In addition to meso-scale robots that move on fabric, there are a number of systems developed to move directly on human skin. SkinBot is a later, tethered version of Rovables, that can hold on to the body using small suction cups and is suitable for tactile feedback applications ~\cite{dementyev2016rovables, dementyev2017skinbot, kao2017exploring, dementyev2018epidermal, dementyev2019dynamic, dementyev2021mechanical}. Dobbelstein~\textit{et al.}~\cite{dobbelstein2018movelet} developed Movelet, a tethered robot that can move along the user's forearm to convey haptic feedback via its movement and positioning. This robot has four wheels for locomotion, weighs 403 g, and its characteristic length (width) is $\sim45$mm.
These studies are supported by Jiang~\textit{et al.}~\cite{jiang2019survey}'s design guidelines for on-body companion robots and Urbani~\textit{et al.}~\cite{urbani2018exploring}'s proposed framework that enables developing experiences to combine AR and wearable companion robots. Furthermore, as the field grows, different application areas for small-scale robots will become available. For instance, Choi~\textit{et al.}~\cite{choi2020bodyprinter} presented a wearable printing device that can print flexible electronics directly on the body. However, this wearable printing device has a limited workspace due to lack of locomotion capabilities.

\subsection{Robot-Assisted Bed Bathing}
Bed bathing has been a significant part of nursing care. Bathing requires a greater focus on safe interaction since washing the body involves direct physical contact with the human body~\cite{8276623}. Current assistance is provided by human caregivers for bathing and hygiene~\cite{cohen2007bathing, goodrich2008human, rader2006bathing}. But, this results in a significant physical and time burden on human caregivers, many of whom are informal non-paid caregivers. 

A number of approaches have been proposed for robot-assisted bathing~\cite{zlatintsi2020support,Erickson2019dressing,5649101,tsumaki2008development,Satoh2009bathingcare,bezerra2017bath,Dometios2018Wiping, 8461568}. 
Tsumaki~\textit{et al.}~\cite{tsumaki2008development} developed a skincare robot that is able to apply ointment to the human's back. King~\textit{et al.}~\cite{5649101} designed a bed bathing system that allows a robot with a compliant arm to perform wiping motions on the surface of a human's arm and leg. Bezerra~\textit{et al.}~\cite{bezerra2017bath} presented a portable washing system to assist in bathing bedridden patients. Zlatintsi and Dometios~\textit{et al.}~\cite{zlatintsi2020support,Dometios2018Wiping,8461568} developed a vision-based washing system that is capable of adapting the motion of a robotic end-effector to a static or moving surface and taking washing actions over the back region of a person. Erickson~\textit{et al.}~\cite{Erickson2019dressing} developed a capacitive sensing approach to track human motion, enabling a mobile manipulator to assist with bathing tasks.

\section{Robot Design and Manufacturing}
\label{sec:design}

\subsection {Design Considerations}
The following design criteria were observed while developing our autonomous cleaning robot. 

\begin{itemize}
\item \textbf{Robot Structure}~
The size and weight of the robot should be small and light enough to achieve efficient and comfortable interactions with the human body. A characteristic length of $\sim 10$cm is desirable for increased comfort and accessibility. In addition, weight of the robot should not exceed $100$g to minimize discomfort on the patient. 
\item \textbf{Safe Human-Robot Interaction}~
The robot should have flexible and soft components to allow for comfortable contact with low applied pressure as it traverses on human skin. The surface should have high adaptability to achieve soft contact and safe interactions with the body.
\item \textbf{Mobility}~
The robot should be able to move on human limbs as well as the torso. This requires adaptibility and modularity of certain components, e.g., flexible materials that can adapt to the changing diameter of the limbs or the ability to move on both flat and cylindrical (convex) surfaces, such as torsos and limbs. 
\item \textbf{Accessibility}~
In order for the robot to be accessible, it should be low-cost and the parts should be easily sourced and replaceable. 3D printing and off-the-shelf components will help to keep the total cost of the robot affordable. In addition, if a robot component is broken, easy replacement will help to increase the lifespan of the robot. 
\end{itemize}

\subsection{Design, Actuation and Fabrication}
\label{sec:methods_subsec}

\begin{figure*}
\centering
\includegraphics[width=0.9\textwidth, trim={0cm 0cm 0cm 0cm}, clip]{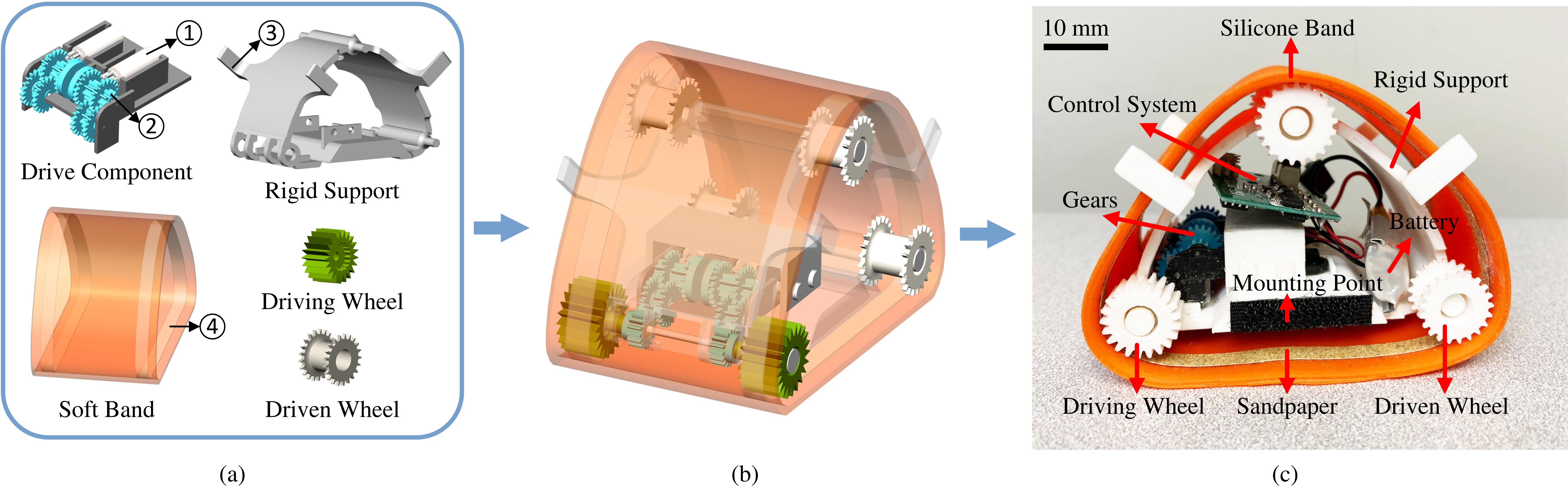}
\caption{\label{fig:3DModel} Mechanical components, assembly, and the physical structure of the robot. 
(a) Main components consist of 1–Motor, 2–Gear train (drive component), 3–Guide Rail on the rigid chassis, and 4–Sandpaper implemented on the soft band. In addition, two sets of wheels (driving - active and driven - passive) provides the rotational motion. (b) 3D CAD model shows the assembled robot. (c) Prototype of the robot. Mounting points are used for connecting cleaning fabric.}
\end{figure*}

The mechanical structure of our cleaning robot is shown in Fig.~\ref{fig:3DModel}. The robot consists of a drive component, a rigid supporting structure, two powered driving wheels, four passive wheels, a soft deformable silicone band, and mounting points for attaching the cleaning cloth. The rigid support structure is the chassis of our robot that houses most of the mechanical and electrical components, i.e., two actuators, drive gear train which transmit the motion to two driving wheels, and control circuit. The triangular cross-section of the chassis provides an optimized contact surface for minimal friction and while providing enough traction. The driving wheel pair is placed at one corner of the triangle and the two pairs of passive wheels are located at the other two corners of the structure. Having only three pairs of wheels is also beneficial to minimize the friction between them and the soft silicone band, that fits around the rigid support. 

\begin{table}
\centering
    \caption{Cleaning Robot Parameters}
        \begin{threeparttable}
        \begin{tabular}{ccc}
            \toprule
            \textbf{Component}&\textbf{Symbol}&\textbf{Value}\\
            \midrule
            \multirow{2}*{\textbf{Rigid Support}}& Weight &14.2 g\\
                                                 & Size &$62.4\times56.0\times42.0$ $(\rm{mm})$\\
            \midrule
            \multirow{2}*{\textbf{Silicone Band}}& Weight & 19.7 g\\
                                                 & Size & $50.8\times63.5\times1.8$ $(\rm{mm})$\\                    
            \midrule                                    
            \multirow{3}*{\textbf{Driving Wheel}}& Weight & 1.0 g\\
                                                 & Length & 8.5 mm\\
                                                 & Maximum Diameter & 15.0 mm\\
            \midrule
            \multirow{3}*{\textbf{Passive Wheel}}& Weight & 0.6 g\\
                                                & Length & 10.0 mm\\
                                                 & Maximum Diameter & 14.0 mm\\
            \midrule
            \multirow{5}*{\textbf{Motor}}& Weight & 1.5 g\\
                                        & Size & $12.0\times6.0$ (mm)\\
                                        & Speed & No-load 45000 RPM\\                                    
                                        & Voltage & DC 3 V\\
                                        & Current & 45 mA\\
            \midrule
            \multirow{2}*{\textbf{Lipo Battery}}& Weight & 3.5 g\\
                                            & Capacity & 100mAh\\
            \bottomrule
            \end{tabular}
            The dimensions and total weight of the robot are $72\times50\times57$ (mm) and 63.4 g, respectively.
        \end{threeparttable}
        \label{table:params}
\end{table}

We use two 612 DC 3 V coreless micro motors to power the robot. Each motor has a maximum speed of $45000$ RPM prior to the gear train. The gear train for one motor consists of six spur gears with a gear ratio of $125/3$, providing no load speed of $1080$ RPM for the drive axle. The design of the gear train is symmetrical, as shown in Fig.~\ref{fig:3DModel}(a). Motor specifications are provided in Table~\ref{table:params}. A $3.7$ V $100$ mAh LiPo battery housed inside the rigid support structure provides power to both motors. We added two lines of 3M sandpaper of grade $150$ grit to the inner surface of the band, as shown in Fig.~\ref{fig:3DModel}. This sandpaper increases contact friction between the driving wheels and the rotating silicone band to support greater driving force. The four passive wheels have internal cutouts such that they make contact with the silicone band rather than the sandpaper to reduce frictional forces that oppose motion. 
Table~\ref{table:params} provides the specifications for each of the robot's components.
We fabricate the rigid support structure and wheels using an Original Prusa i3 3D printer with polylactic acid (PLA) material, and the final robot is shown in Fig.~\ref{fig:3DModel}(c). 
The robot is radio controlled (RC) by a transmittor with a signal range of up to 10 meters.



\subsection{Fabric Mounting and Human Interaction}

The cleaning task of robot is achieved by fabric that is attached to both sides of the robot at mounting points. This connection creates a loop so that the robot can clean objects with circular cross-section. The length of the fabric is adjusted to provide contact with the object to be cleaned. 

First, the robot body is placed on the object, then the fabric is placed around the object and secured on the robot with the help of hook-and-loop fastener. When the robot is powered, driving wheels begin to rotate and exert force on the silicone band, resulting in the rotation of silicone band around the chassis. The guides on the chassis are designed to ensure that the band does not slip out of place. As the robot moves forward, it also pulls the fabric along the object, which provides cleaning.  


A cleaning action of the robot refers to pushing away colored particles on the object as the robot moves. This pushing action is achieved due to the close contact of the fabric with the object. In our experiments, we evaluate multiple fabric types according to different elasticity and friction properties. 



\subsection{Simplifying Assumptions}

The robot's cleaning ability is evaluated on an object, i.e. PVC pipe, with a constant diameter. In order to understand the performance of the robot on objects with varying diameter, i.e., human limbs, we performed preliminary tests. Using stretchable fabric accommodates the change in object diameters up to a certain point, but additional mechanisms are required for large diameter variations. 

In order to ensure that the robot does not slip off of the smooth pipe, a clear double-sided tape was adhered on top of the pipe. This tape provides a straight path for the robot. However, usage of tape was not necessary for experiments on human limbs as the robot did not exhibit any slipping behavior.  



\section{Experiments}
\label{sec:experiments}

The cleaning performance of our robot is evaluated by a set of experiments based on the fabric parameters and the experimental setup. Four different types of fabric at three widths are used for the cleaning tasks. In order to understand the effect of the fabric on the friction, robot speed is calculated. We use image processing techniques to quantitatively analyze the results on three different surfaces: (i) constant diameter smooth pipe, (ii) varying diameter human limb, and (iii) manikin torso.


\begin{figure}
\centering
\includegraphics[width=0.45\textwidth, trim={0cm 0cm 0cm 0cm}, clip]{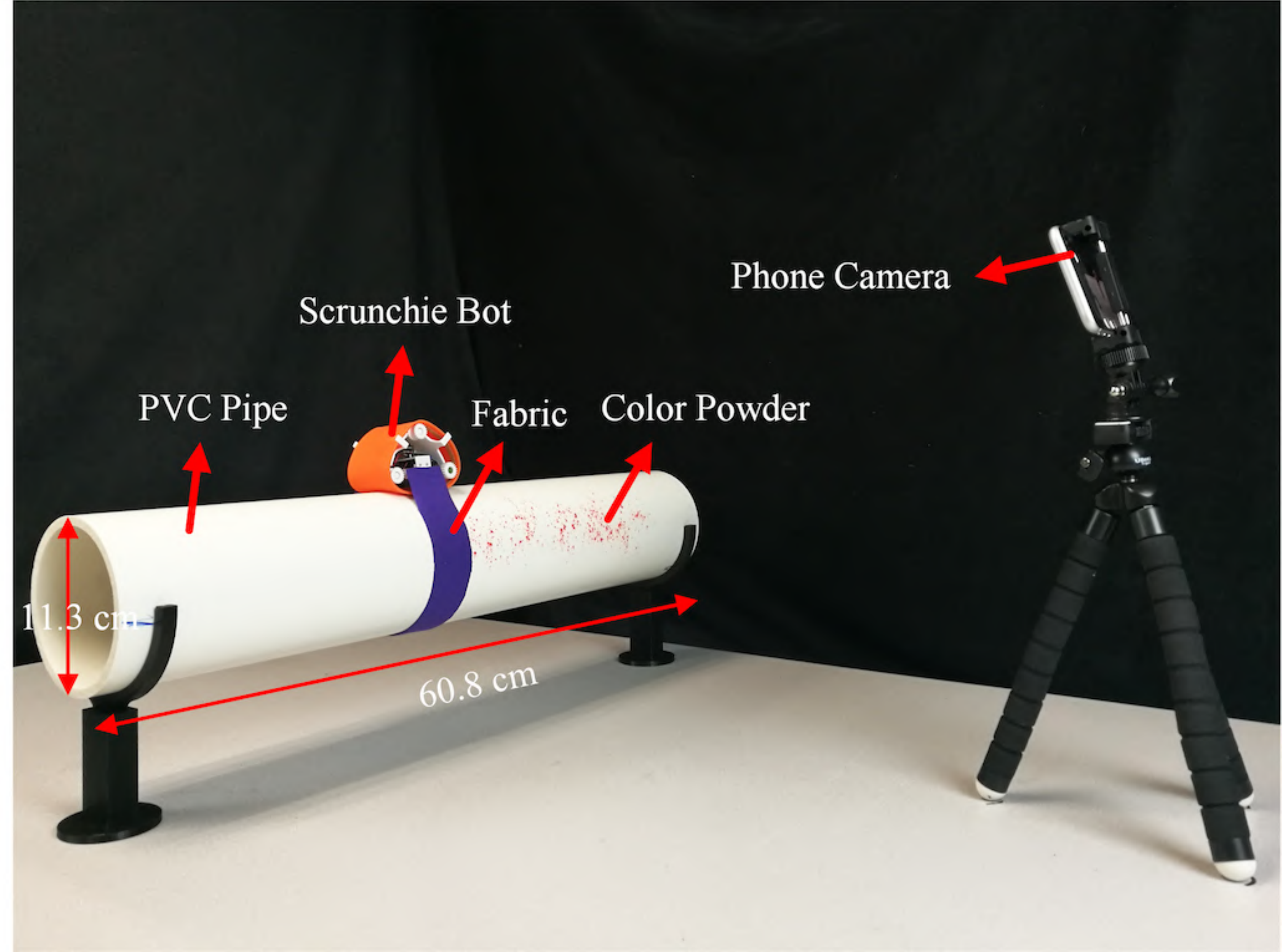}
\caption{\label{fig:plateform} Experimental setup. The robot is mounted on the pipe and moves along the top with the fabric cleaning the debris (i.e., red color powder). }
\end{figure}

\subsection{Experimental Setup}

Our experimental setup consists of an object to be cleaned and a camera to record the cleaning process as shown in Fig.~\ref{fig:plateform}. The cleaning task was performed on three different media. First, a PVC pipe with a diameter of $11.3$ cm and a length of $60.8$ cm is used for evaluation. For this task, the pipe is cleaned with disinfecting wipes and then dried with a dry fabric. A Holi colored corn starch powder is spread over the pipe randomly. Since the robot and setup is symmetrical, only one side of the pipe is recorded. Three photos were taken for analysis: one before distributing the colored powder, one before the cleaning process and one after the cleaning process, and they are successively compared to compute the success rate. 

The second surface that the robot performed a cleaning task with was the arm of a research experimenter, that was positioned on the black support structures for stability. An additional demonstration is performed on a manikin torso. In order to clean the flat surface of a human torso, a fabric layer was adhered on the soft silicone band, enabling the robot to clean the front and back torso of the manikin while moving on them. For the experiments on the pipe and human arm, the fabric is dry and the colored powder is pushed away from the surface as the robot moves. However, for the torso surface, the fabric was sprayed with water for colored powder to stick.




\subsection{Evaluation Method}

\begin{figure}
\centering
\includegraphics[width=0.45\textwidth, trim={0cm 0cm 0cm 0cm}, clip]{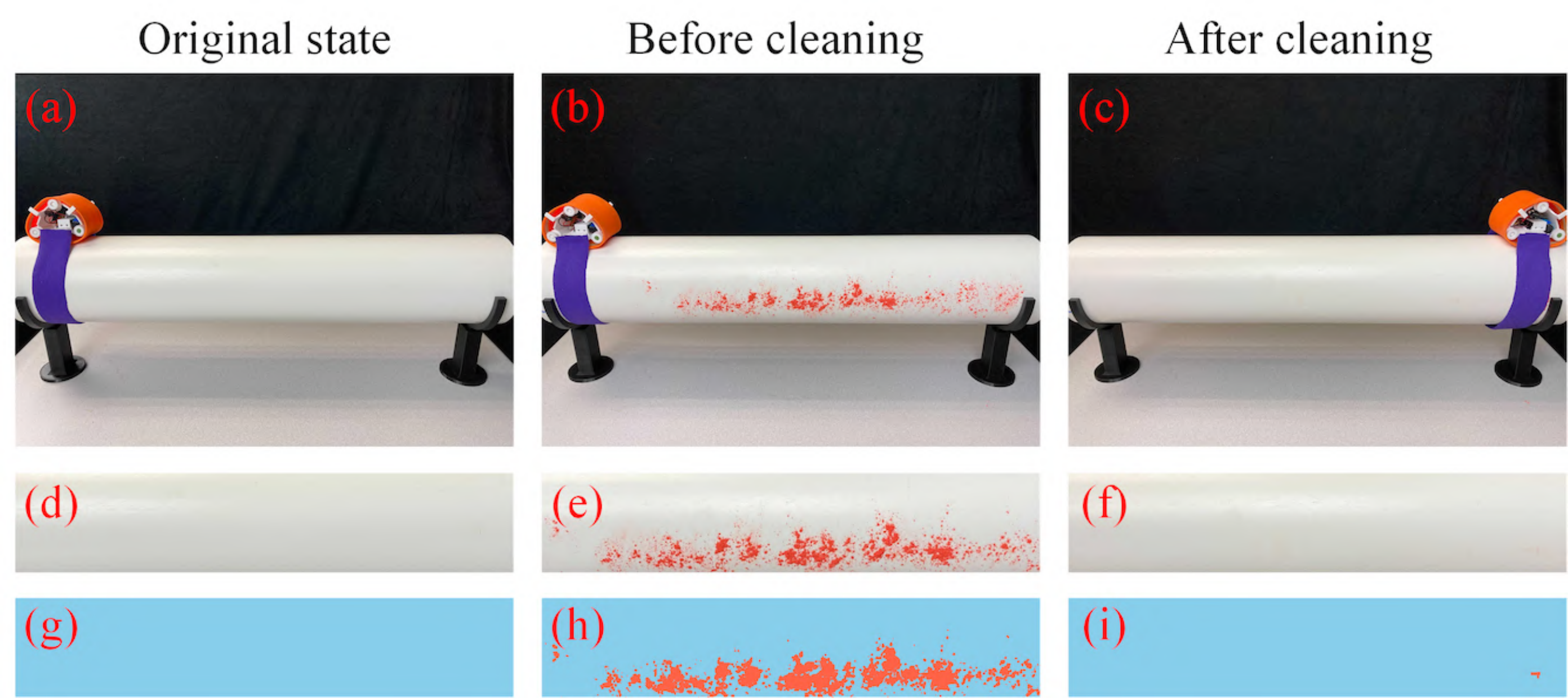}
\caption{\label{fig:method} Evaluation of the cleaning success rate using image processing. (a), (b), and (c) are images of original state (before the random distribution of colored powder), before, and after cleaning, respectively. (d), (e) and (f) are images after projection transformation. (g), (h) and (i) are binary images.}
\end{figure}

In order to characterize the cleaning performance of our robot, we use image binarization to compare the surface before and after the cleaning task, as shown in Fig.~\ref{fig:method}. First, we crop images to remove the background and focus on the pipe surface to be cleaned. The pixel resolution of these images are $1500\times300$. From this, we obtain corresponding grayscale images. We then compute binary images highlighting the colored powder on the pipe via color thresholding. The success rate of cleaning is calculated as,


\begin{equation}
\begin{aligned}
& \begin{split}
P = \frac{A_b-A_a}{A_b-A_o},
\end{split}
\end{aligned}
\end{equation}

where $A_o$, $A_b$, and $A_a$ are the areas of the red pixels (debris pixels) in the original, before, and after cleaning images, respectively.

\begin{figure}[hb]
\centering
\includegraphics[width=0.45\textwidth, trim={0cm 0cm 0cm 0cm}, clip]{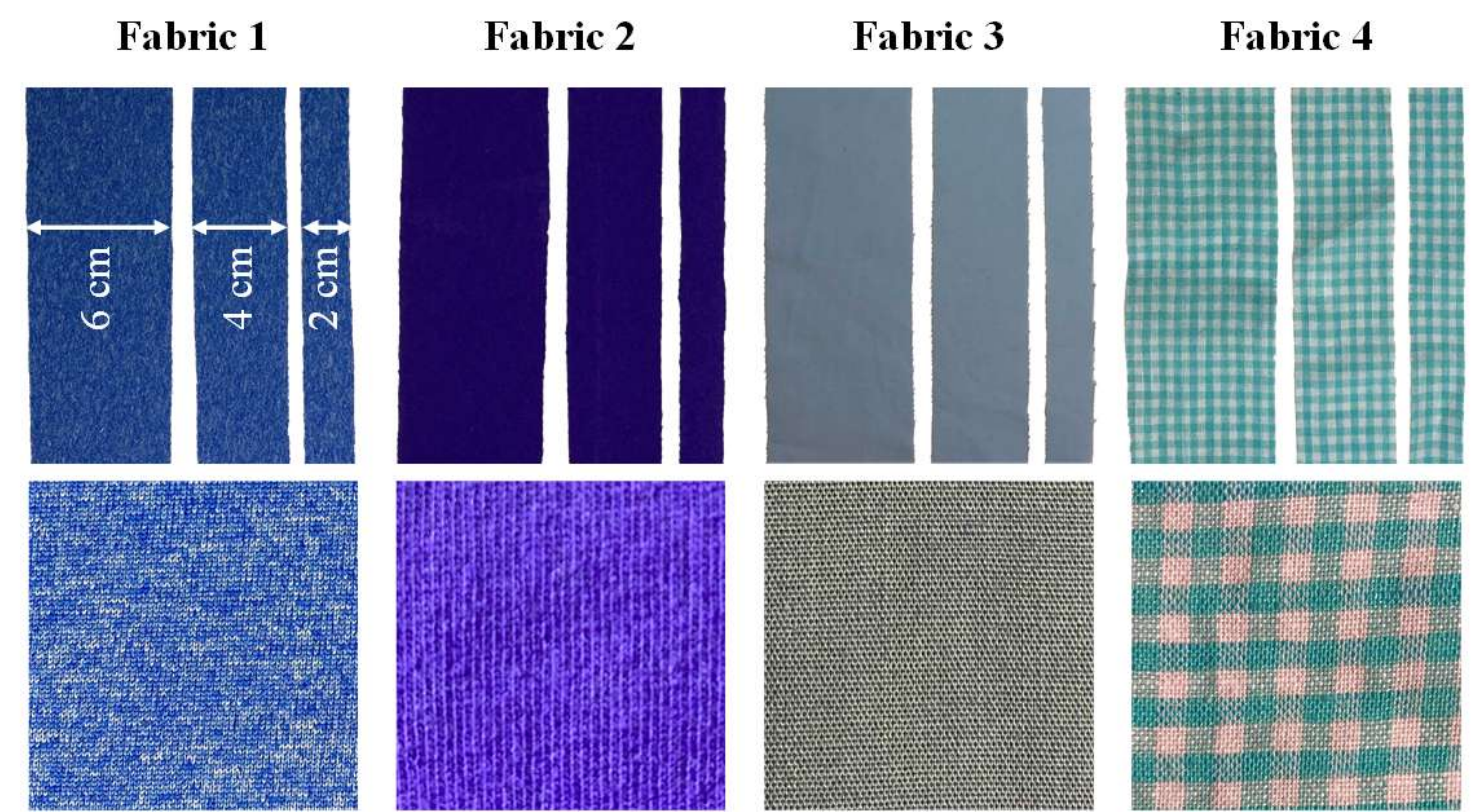}
\caption{\label{fig:fabric} Four types of fabric are used for cleaning. Fabric 1—100\% Polyester, Fabric 2—100\% Cotton, Fabric 3—55\% Cotton and 45\% Polyester, Fabric 4—45\% Cotton and 55\% Linen. Fabric 1 is most elastic among the four, Fabric 2 is the second. most elastic. Fabrics 3 and 4 do not exhibit stretch. We performed experiments with three different widths of fabric; $6$cm, $4$cm, and $2$cm.}
\end{figure}

\begin{figure*}
\centering
\includegraphics[width=0.9\textwidth, trim={0cm 32.5cm 0cm 0cm}, clip]{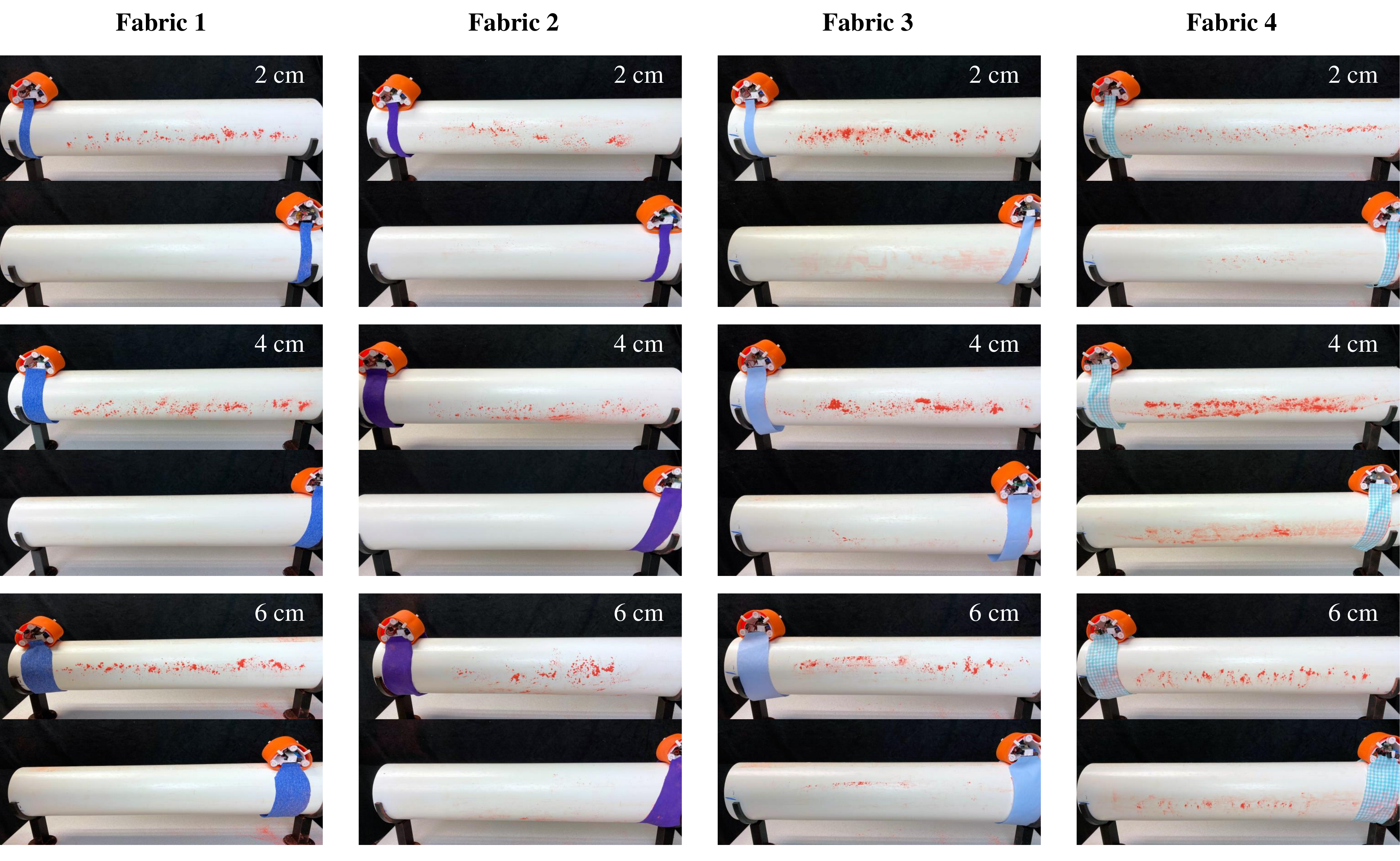}
\caption{\label{fig:fabrictest} Cleaning performances with four types of fabrics. Snapshots of before and after cleaning with different fabrics (columns) and fabric widths (rows).}
\end{figure*}

\subsection{Fabric Evaluation}
\label{sec:fabric_eval}
An important component of our robot is the fabric that is attached to it as it effects the robot's motion and the cleaning performance. In order to find the most suitable fabric used for cleaning, we performed the cleaning experiments on a PVC pipe with four types of fabrics having three different widths, as shown in Fig.~\ref{fig:fabric}. These materials can be classified into two categories: elastic (fabrics 1 and 2) and inelastic (fabrics 3 and 4). All of the experiments are repeated five times. 

Fig.~\ref{fig:fabrictestplots} shows the relationship between the cleaning success rate $P$ and the types of fabric. The experiments demonstrate that the two types of elastic fabrics, i.e., fabrics 1 and 2 perform noticeably better than the other two inelastic fabrics, i.e., fabrics 3 and 4. This is due to the sufficient contact between the elastic fabrics and pipe surface, enabling easier debris removal. The average cleaning percentage remains above $90$\% for fabrics 1 and 2. All of the success rates and standard deviations are presented in Fig.~\ref{fig:fabrictestplots}~(a). Moreover, fabric strips with a width of $4$ cm had the best performance except for fabric 4, indicating that the fabric width is a key factor in cleaning effectiveness.

\begin{figure*}[t]
\centering
\includegraphics[width=0.9\textwidth, trim={0cm 0cm 0cm 34cm}, clip]{Figure7.pdf}
\caption{\label{fig:fabrictestplots} Cleaning evaluations with four types of fabrics depicted in Fig. 6. (a) Average success rate with respect of fabric types and (b) the effect of fabric type and width on the moving velocity of the robot.}
\end{figure*}

\begin{figure}
\centering
\includegraphics[width=0.45\textwidth, trim={0cm 0.5cm 0cm 0cm}, clip]{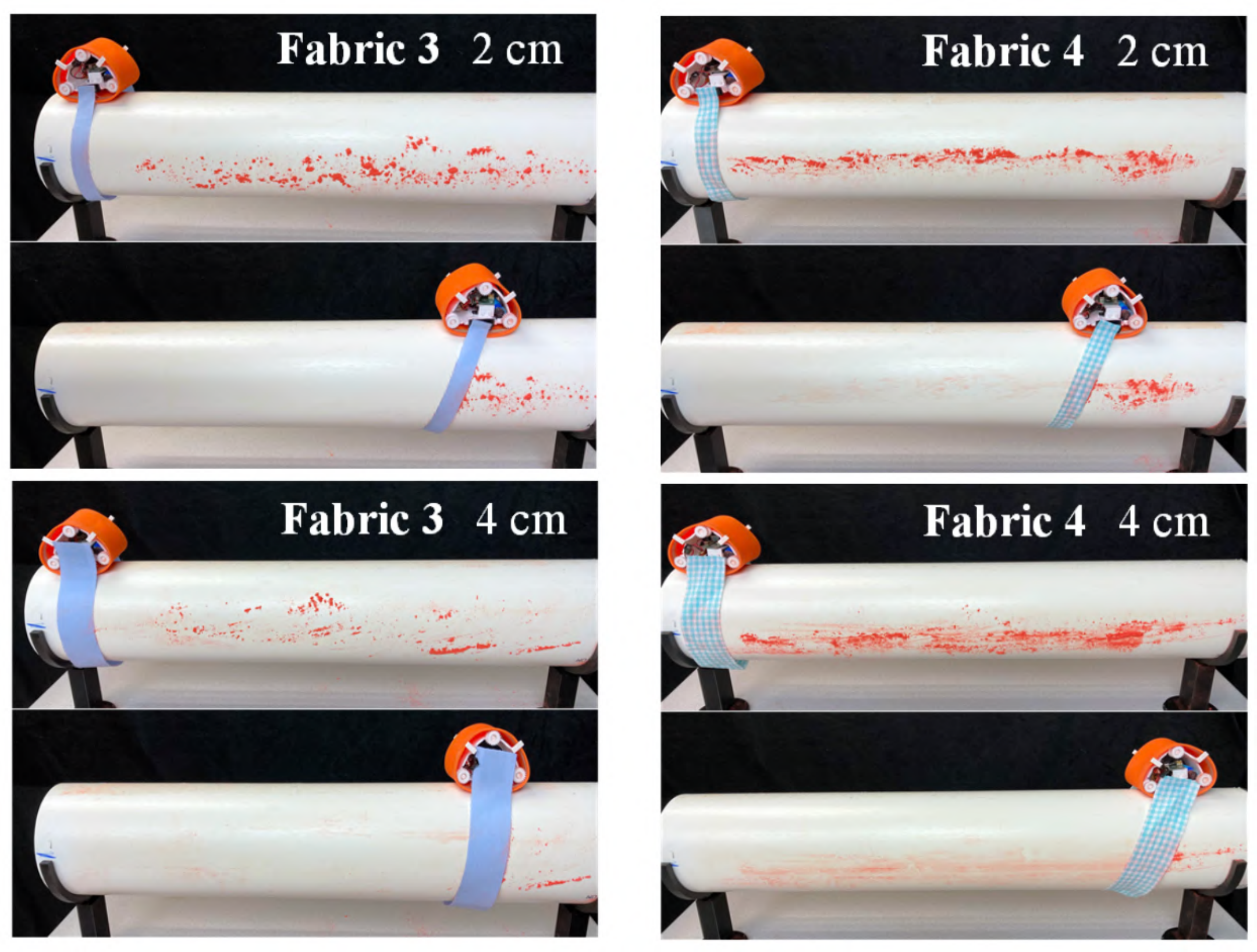}
\caption{\label{fig:fabrictestfails} Examples of the robot getting stuck for unstretchable fabrics 3 and 4.}
\end{figure}

\begin{figure*}
\centering
\includegraphics[width=0.9\textwidth, trim={0cm 0cm 0cm 0cm}, clip]{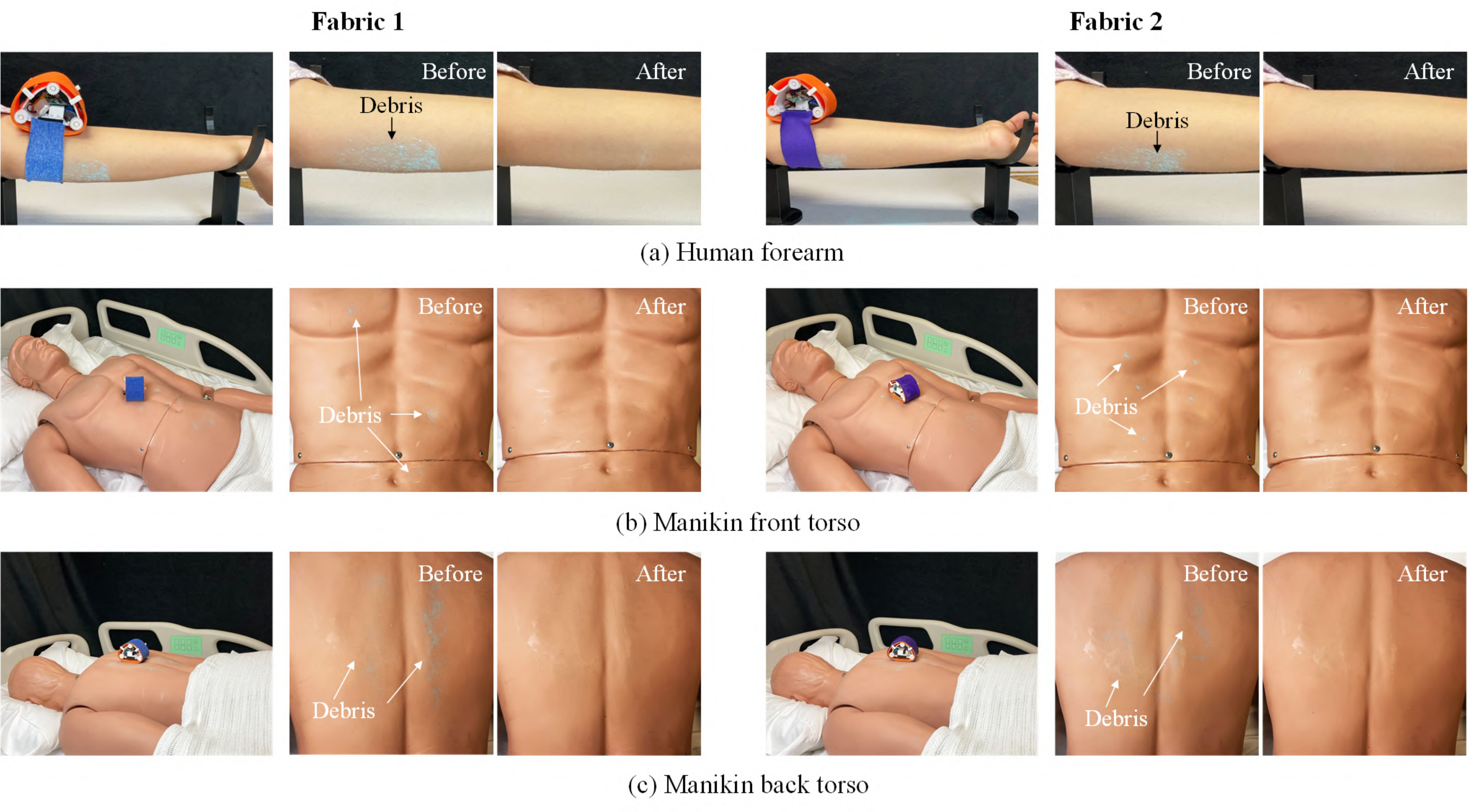}
\caption{\label{fig:bodytest} Snapshots before and after cleaning tasks. (a) Experiments on a research experimenter's forearm. (b) Experiments on manikin front torso. (c) Experiments on manikin back torso.}
\end{figure*}

\begin{figure}
\centering
\includegraphics[width=0.45\textwidth, trim={0cm 0cm 0cm 0cm}, clip]{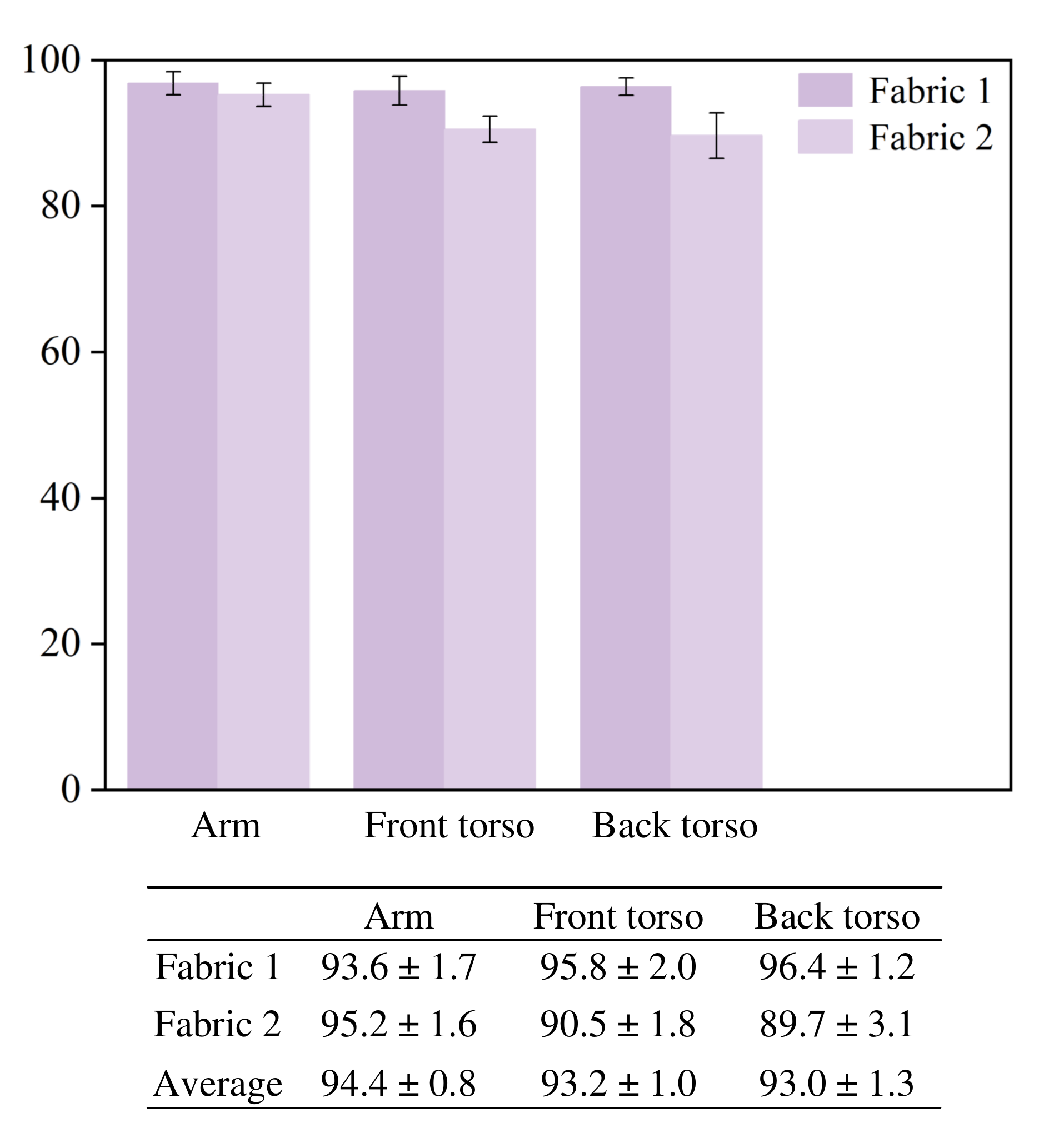}
\caption{\label{fig:bodypercentage} Cleaning percentage (\%) for scenarios depicted in Fig.~\ref{fig:bodytest}.}
\end{figure}

Fig.~\ref{fig:fabrictestplots}~(b) depicts the average moving speed of the robot across all of the trials for each fabric type and width. In all trials, the robot received the same motor control commands to move along the pipe at a fixed velocity. Overall, we observed that wider fabric reduced kinetic friction for all fabric types, which allowed the robot to move at higher translational speeds.

Cleaning performance can be improved slightly with the two inelastic fabrics by mounting them more tightly around the pipe, as shown in Fig.~\ref{fig:fabrictestfails}. However, doing so increased surface friction, which in turn reduced the robot's translational speed along the pipe and resulted in the robot getting stuck along the pipe due to high kinetic friction.




\subsection{Wiping Behavior on the Human Body}

In order to evaluate the cleaning performance of the robot on the human body, we performed six groups of experiments on the arm of a human experimenter and on a manikin torso. Each experiment was repeated five times. 


Since the robot achieved the best cleaning progress on the PVC pipe with the $4$cm wide elastic fabrics, the experiments on the human arm is performed by this choice. 

We performed the cleaning task on the arm of a human experimenter, as shown in Fig.~\ref{fig:bodytest}~(a). The robot used the two $4$cm wide elastic fabrics as these achieved the best cleaning progress on the PVC pipe (see Section~\ref{sec:fabric_eval}). We placed blue colored powder on the surface of the limb for the purpose of easily distinguishing the debris from the color of the skin. The arm was supported by two holders to remain stationary. The robot moved along the forearm to wipe off the debris. In addition, we attached the elastic fabrics onto the outside of the soft silicone band to test the cleaning ability of the robot on a manikin torso, as shown in Fig.~\ref{fig:bodytest}~(b) and (c). During these experiments, we spread blue colored powder on both the front and back torso of the manikin. We kept the fabric damp (slightly wet) for the purpose of adhering powder to the surface of the fabric, similar to a washcloth used for bathing. We robot moved along the torso in a linear path, following 6 trajectories from different directions to clean the torso. 


The results of the cleaning tasks on the forearm and manikin torso are shown in Fig.~\ref{fig:bodypercentage}. Our robot with elastic fabrics successfully cleaned on average 94.4\% of the visible powder on the forearm, 93.2\% on the manikin's chest, and 93.0\% of the visible powder on the manikin's back. These experiments indicate that a soft wearable robot that traverses along the body remains a promising research direction for bathing and skin care.

\section{Conclusion}
\label{sec:conclusion}

In this work, we introduce a meso-scale robot that locomotes along the human body to provide bathing and skin care assistance. The small size and light weight of the robot supports its use as an assistive robot that moves along the body. The soft surface of the robot provides high adaptibility to the surface it is moving on and safe contact with the human skin. The characterization tests to evaluate the success rate of cleaning performance are completed on a cylindrical pipe using four different types of fabric. The results show that elastic fabrics achieve the highest success rate by cleaning more than $91\%$ of the debris from the pipe surface. In addition, we conduct evaluations in which the robot cleans off the surface of a human experimenter's forearm and of a medical manikin's torso---successfully cleaning off greater than 93\% of all debris across all trials.





\bibliographystyle{IEEEtran}
\bibliography{bibliography}


\end{document}